\documentclass[aic,crcready]{iosart2x}
\usepackage{verbatim}

\pubyear{0000}
\volume{0}
\firstpage{1}
\lastpage{1}

\begin{document}

\begin{frontmatter}

\title{Neuro-Symbolic Artificial Intelligence}
\runtitle{Neuro-Symbolic Artificial Intelligence: Current Trends}
\subtitle{Current Trends}


\author
{\inits{Md. K.}\fnms{Md Kamruzzaman} \snm{Saker}\ead[label=e2]{second@somewhere.com}},
\author
{\inits{L.}\fnms{Lu} \snm{Zhou}\ead[label=e3]{third@somewhere.com}},
\author
{\inits{A.}\fnms{Aaron} \snm{Eberhart}\ead[label=e2]{second@somewhere.com}},
and
\author
{\inits{P..}\fnms{Pascal} \snm{Hitzler}\ead[label=e1]{hitzler@ksu.edu}%
\thanks{Corresponding author. \printead{e1}.}}
\address
{Department of Computer Science, \orgname{Kansas State University},
KS, \cny{USA}\printead[presep={\\}]{e1}}
\runauthor{Md K. Sarker, L. Zhou, A. Eberhart, P. Hitzler}

\begin{abstract}
Neuro-Symbolic Artificial Intelligence -- the combination of symbolic methods with methods that are based on artificial neural networks -- has a long-standing history. In this article, we provide a structured overview of current trends, by means of categorizing recent publications from key conferences. The article is meant to serve as a convenient starting point for research on the general topic.
\end{abstract}


\end{frontmatter}


\section{What Is Neuro-Symbolic Artificial Intelligence?}\label{sec:intro}

We should begin by attempting to define the subject matter. In a general sense, we understand \emph{Neuro-Symbolic Artificial Intelligence} (in short, \emph{NeSy AI}), to be a subfield of the field of Artificial Intelligence (in short, \emph{AI}), which focuses on bringing together, for added value, the \emph{neural} and the \emph{symbolic} traditions in AI. Different spellings are currently in use, that include \emph{neural-symbolic} and \emph{neurosymbolic}, but also \emph{symbolic-subsymbolic} and others -- which we consider to be equal. The term \emph{neural} in this case refers to the use of artificial neural networks, or connectionist systems, in the widest sense. The term \emph{symbolic} refers to AI approaches that are based on explicit symbol manipulation. This in general would include things like term rewriting, graph algorithms, and natural language question answering. It is often more narrowly understood, though, as a reference to methods based on formal logic, as utilized, for instance, in the subfield of AI called Knowledge Representation and Reasoning. The lines easily blur, though, and for the purposes of this overview, we will not restrict ourselves to logic-based methods only.

The general promise of NeSy AI lies in the hopes of a best-of-both worlds scenario, where the complementary strengths of neural and symbolic approaches can be combined in a favorable way. On the neural side, the desirable strengths would include trainability from raw data and robustness against faults in the underlying data, while on the symbolic side one would like to retain the inherently high explainability and provable correctness of these systems, as well as the ease of making use of deep human expert knowledge in their design and function. In terms of functional features, utilizing symbolic approaches in conjunction with machine learning -- in particular with deep learning, which is currently most busily researched on -- one would hope to do better on issues like out of vocabulary handling, training from small data sets, recovery from errors, and in general, explainability, as opposed to systems that rely on deep learning alone.

One of the fundamental differences between neural and symbolic AI approaches, that is relevant for our discussion, is that of representation of information within an AI system. For symbolic systems, representation is explicit and in such terms that are in principle understandable by a human. E.g., a rule such as $\text{square}(x) \to \text{rectangle}(x)$ is readily understood and manipulated by symbolic means. In neural systems, though, representations are usually by means of weighted connections between (many) neurons and/or simultaneous activations over a (possibly large) number of neurons. In particular, a human observer would not be able to readily recognize what is being represented. In a deep learning context, these distributed representations are called \emph{embeddings}, are learned during training, and are thus an explicit part of the architecture of the deep learning system. On a fundamental level, these two types of representing information are very different indeed. Representations by vectors of real numbers -- or even more ephemerally, by connections weighted by real numbers -- are in terms of entities within a high-dimensional and continuous, differentiable vector space as required for standard learning algorithms, and are sometimes called sub-symbolic representations. Symbolic representations, on the other hand, are by their nature discrete and their natural relationships (such as implications) do not readily lend themselves to representation into a continuous space.\footnote{This gap between the discrete and the continuous can be bridged by mathematical means, e.g. using Cantor Space as in \cite{HitzlerS03}. However the approach did not scale.} Correspondingly, representational questions are often at the heart of the development of neuro-symbolic approaches.

Neuro-Symbolic AI has a long-standing history, with even deeper roots, such as the McCulloch \& Pitts 1943 landmark paper \cite{MP43}, that pre-date the field of AI. A lot of work on the topic is also not explicitly placed as NeSy AI research, as the term itself only begun to be used more systematically since the 1990, and only very recently -- as we will see below -- has seen an increase in the context of major AI conferences. The general theme is also of importance in the context of the field of Cognitive Science \cite{Besold17survey}. 

We know of one series of annual workshops on the topic -- the International Workshop on Neural-Symbolic Learning and Reasoning\footnote{See \url{http://neural-symbolic.org/}.} -- that dates back to 2005. But in recent years, several additional venues have been established, including the AAAI Spring Symposium AAAI-MAKE on Combining Machine Learning and Knowledge Engineering,\footnote{\url{https://www.aaai-make.info/}} the workshop on Combining Symbolic and Sub-Symbolic Methods and Their Applications (CSSA),\footnote{\url{https://alammehwish.github.io/cssa_cikm_2020/}} the workshop on Cognitive Computation: Integrating neural and symbolic approaches, CoCo,\footnote{http://neural-symbolic.org/} and the Expo Workshop on Perspectives on Neurosymbolic Artificial Intelligence Research\footnote{\url{https://nips.cc/Conferences/2020/Schedule?showEvent=20241}} as well as the workshop on Knowledge Representation \& Reasoning Meets Machine Learning, KR2ML,\footnote{\url{https://kr2ml.github.io/2020/}} both at NeurIPS 2020. We will herein focus mostly on recent developments, from the past few years. For the reader interested in getting an overview of previous efforts, we would like to refer to past overview works such as \cite{Besold17survey, 2005-nesy-survey, nesy-book-07,GarcezLG2009}. 

The remainder of this article will focus on providing an overview of recent research contributions to the NeSy AI topic as reflected in the proceedings of leading AI conferences. In order to provide a structured approach to this, we will group these recent papers in terms of a topic categorization proposed by Henry Kautz at an AAAI 2020 address. We will also categorize the same recent papers according to a 2005 categorization proposal \cite{2005-nesy-survey}, and discuss and contrast the two. In Section \ref{sec:category} we will first present the two categorizations. In Section \ref{sec:past} we will then discuss the categorization results, and in Section \ref{sec:future} we will provide a future outlook.

\section{Categorizing Neuro-Symbolic Artificial Intelligence}\label{sec:category}

Henry Kautz, in his AAAI 2020 Robert S. Engelmore Memorial Award Lecture, discusses five categories for neuro-symbolic AI systems in his discussion about the "Future of AI":\footnote{\url{https://www.youtube.com/watch?v=_cQITY0SPiw}, starting at minute 29}
\begin{description}
\item[\mbox{[symbolic Neuro symbolic]}] refers to an approach where input and output are presented in symbolic form, however all the actual processing is neural. This, in his words, is the "standard operating procedure" whenever inputs and outputs are symbolic.
\item[\mbox{[Symbolic[Neuro]]}] refers to a neural pattern recognition subroutine within a symbolic problem solver, with examples such as AlphaGo, AlphaZero, and current approaches to self-driving cars.
\item[\mbox{[Neuro $\cup$ compile(Symbolic)]}] refers to an approach where symbolic rules are "compiled" away during training, e.g. like the 2020 work on Deep Learning For Symbolic Mathematics \cite{DBLP:conf/iclr/LampleC20}.
\item[\mbox{[Neuro $\rightarrow$ Symbolic]}] refers to a cascading from a neural system into a symbolic reasoner, such as in the Neuro-Symbolic Concept-Learner \cite{DBLP:conf/iclr/MaoGKTW19}.
\item[\mbox{[Neuro[Symbolic]]}] refers to the embedding of symbolic reasoning inside a neural engine, where symbolic reasoning is understood as "deliberative, type 2 reasoning" as common, e.g., in business AI, and including an internal model of the system's state of attention: Concepts are decoded to symbolic entities in an attention schema when attention to these concepts is high. A goal in the attention schema then signals that deliberative reasoning be executed. 
\end{description}
Kautz presented these as "five ways to bring together the neural and symbolic traditions." In the brief address, only a few examples were given of corresponding work. 

In a much older survey from 2005 \cite{2005-nesy-survey}, work on NeSy AI was classified according to eight dimensions, grouped into three aspects:
\begin{itemize}
    \item Interrelation
    \begin{enumerate}
        \item \textbf{Integrated versus hybrid} -- this was included mostly for delineation to hybrid systems; indeed the focus was entirely on integrated systems, where symbolic processing functionalities emerge from neural structures and processes. We will have the same emphasis herein.
        \item \textbf{Neuronal versus connectionist} -- this aspect refers to the question whether the artificial neural network architecture is dominantly chosen in order to mimic biological neural networks (neuronal) versus a more technological approach that focused on desireable computational properties as emphasized in the field of AI (connectionist).
        \item \textbf{Local versus distributed} -- this dimension refers to the question how symbolic information is represented within the neural system, namely whether this happens in a very localist way (one or a few dedicated neurons for a symbolic piece of knowledge) or in a highly distributed way, where a representation consists of a larger number of activation values and/or weights.
        \item \textbf{Standard versus nonstandard} -- this refers to the question whether a NeSy AI system draws on artificial neural network architectures that are \emph{standard} in the sense that they are widely used, propagate real numbers and nodes compute only simple functions.
    \end{enumerate}
    \item Language
    \begin{enumerate}\setcounter{enumi}{4}
        \item \textbf{Symbolic versus logical} -- this refers to the symbolic aspect of the NeSy AI system, namely whether it is based on formal logic (logical) or on other symbolic processing (such as structured datatypes like graphs).
        \item \textbf{Propositional versus first-order} -- for the logical aspect only, a delineation whether the underlying logic is propositional in nature, or more expressive like using higher arity predicates and quantifiers.
    \end{enumerate}
    \item Usage
    \begin{enumerate}\setcounter{enumi}{6}
        \item \textbf{Extraction versus representation} -- this aspect refers to the information flow within the system, namely whether it primarily \emph{extracts} symbolic information from a neural substrate, or whether it is primarily based on neural representations of symbolic knowledge.
        \item \textbf{Learning versus reasoning} -- this refers to the core functionality of the system, namely whether its focus is on machine learning or on automated symbolic reasoning.
    \end{enumerate}
\end{itemize}

These eight dimensions presented a view of the existing facets of the field in 2005, and examples were given for each of the dimensions. It is important to note that they were presented as dimensions, and not as binary values, e.g., a system may fall anywhere on a continuum or even fall under both aspects of the dimension (i.e., span the dimension). We gather that the above listed dimensions are mostly self-explanatory as described; further details can be found in \cite{2005-nesy-survey}.

We started out on our investigations with the hypothesis that the NeSy AI field has shifted focus in recent years, e.g. rendering some of the 2005 dimensions obsolete, at least for the time being. We will discuss this in more detail below. At this stage, however, we remark that the Kautz categories listed above seem to address only Interrelation aspects. However they cover these in a rather different way than the 2005 survey, with a focus on more precise architectural description of the system workflows.

\begin{figure}
    \centering
    \includegraphics[width=.45\textwidth]{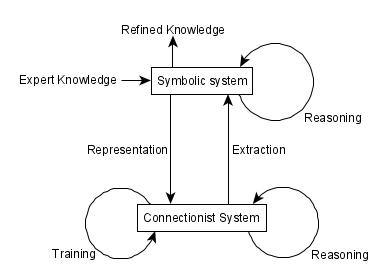}
    \caption{Neuro-symbolic learning cycle according to \cite{2005-nesy-survey}}
    \label{fig:nesy-cycle}
\end{figure}

Interestingly, though, a diagram presented in the 2005 survey (see Figure \ref{fig:nesy-cycle}) bears striking similarity to a diagram from Kautz' presentation,\footnote{at minute 35} and indeed the Kautz categories can be understood as utilizing part of this diagram as architectural substrate, and can serve by itself as a description of the essence of NeSy AI. In particular, in the diagram note that I/O is situated on the symbolic side only, while training is situated only on the neural side, whereas reasoning can happen in either part. This is congruent with the best of both worlds perspective discussed earlier, as neural systems excel in training (and in reasoning despite of faults in the data), whereas symbolic systems excel in dealing with explicit knowledge (or serving human-understandable explanations) and sport provably correct reasoning procedures.

\section{Neuro-Symbolic Artificial Intelligence: Trends}\label{sec:past}

For the purpose of this overview, and as a guide to recent literature, we had a look at mainstream and top tier AI publications on NeSy AI. We particularly looked at all published papers since 2011 (except ICLR which started in 2013), at top conference venues, and selected all original research papers on NeSy AI; we disregarded overview and survey works. The conferences we looked at, were
\begin{itemize}
    \item the International Conference on Machine Learning (ICML),
    \item the Conference and Workshop on Neural Information Processing Systems (NeurIPS formerly NIPS),
    \item the AAAI Conference on Artificial Intelligence (AAAI)
    \item the International Joint Conference on Artificial Intelligence (IJCAI), and
    \item the International Conference on Learning Representations (ICLR).
\end{itemize}
In order to determine whether a paper falls into the NeSy AI theme, we read the abstract (and sometimes the introduction). As mentioned before, we are aware that not all papers relevant for NeSy AI are phrased in such terms, i.e. we acknowledge that we may have missed a few relevant papers. It still seems a reasonable assumption that the sum of our selected papers represents a valid cross-section. We are also aware that restricting our attention to the above-mentioned five conferences leaves out a lot of relevant work. However our focus was on recent, mainstream AI research, and we believe that our selection is reasonable for this purpose. It is conceivable that an analysis of publications at second-tier conferences and at workshops, or in other fields such as Cognitive Science, may provide a different picture. However this was not the focus of our effort. 

The number of papers, per year, for each of the conferences is listed in Table \ref{tab:pap-year-conf}. Several things are already remarkable.
\begin{itemize}
    \item There was only one paper before 2016, and 40 papers since 2017. This alone indicates a shift towards a more mainstream adoption of the NeSy AI research area. 
    \item The total paper count, at these selected conferences, is still rather low compared to the overall number of accepted papers. E.g., IJCAI-PRICAI 2020 had 592 accepted papers, and NeurIPS 2020 had 1900 accepted papers. 
    \item We do not have insights into the topics of rejected papers, but with acceptance rates at these conferences often at 20\% or lower, it seems reasonable to expect that the total number of submitted NeSy AI papers, most of which likely published elsewhere in the meantime, was at least about five times higher.
\end{itemize}

\begin{table}
\begin{tabular}{l|rrrrrrrrrr|r}
conference &  2011 & 2012 & 2013 & 2014 & 2015 & 2016 & 2017 & 2018 & 2019 & 2020 & total \\
\hline
ICML & 0 & 0 & 0 & 0 & 0 & 1 & 3 & 2 & 5 & 6 & 17 \\
NeurIPS & 0 & 0 & 0 & 0 & 0 & 0 & 0 & 4 & 2 & 4 & 10\\
AAAI & 0 & 0 & 0 & 0 & 0 & 1 & 0 & 1& 1 & 1 & 4\\
IJCAI & 1 & 0 & 0 & 0 & 0 & 0 & 2 & 2 & 0 & 1 & 6\\
ICLR & N/A & N/A & 0 & 0 & 0 & 0 & 1 & 1 & 1 & 3 & 6\\
\hline
total & 1 & 0 & 0 & 0 & 0 & 2 & 6 & 10 & 9 & 15 & 43
\end{tabular}
\caption{Included paper counts, per year and conference. }\label{tab:pap-year-conf}
    \end{table}

We now manually assessed each of these papers as to their categorization with respect to the two schemes described in Section \ref{sec:category}. Doing so was not straightforward at all, as the descriptions of the categories are somewhat ambiguous, and it is not always clear which categorization would be most adequate. Still, from the resulting Table~\ref{tab:kautz} we can see that the first four Kautz categories seem to provide a reasonably balanced perspective on the recent publications at these conferences. The fifth category in fact, as also discussed by Kautz in his address, was meant to be more forward-looking, a goal to aspire to in future research. So we are not surprised to not see any papers in this category.

\begin{table}
\begin{tabular}{lrl}
category & number of papers & papers\\
\hline
\mbox{[symbolic Neuro symbolic]} & 13  & \cite{DBLP:conf/ijcai/YangLLG18,DBLP:conf/aaai/DemeterD20,DBLP:conf/aaai/LyuYLG19,DBLP:conf/aaai/CaoLX16,DBLP:conf/nips/ChenLYSZ20,DBLP:conf/nips/XieXMKS19,DBLP:conf/nips/ManhaeveDKDR18,DBLP:conf/iclr/ArabshahiSA18,DBLP:conf/ijcai/PenningGLM11}\\
\mbox{[Symbolic[Neuro]]} & 9  & \cite{DBLP:conf/nips/JiangA20,DBLP:conf/nips/LiangHZLX18,DBLP:conf/nips/ManhaeveDKDR18,DBLP:conf/iclr/CohenSHS20,DBLP:conf/icml/TeruDH20,DBLP:conf/icml/GargBM20,DBLP:conf/icml/MouLLJ17}\\
\mbox{[Neuro $\cup$ compile(Symbolic)]} & 10  & \cite{DBLP:conf/ijcai/XiaoDG17,DBLP:conf/nips/Dang-Nhu20,DBLP:conf/nips/Yi0G0KT18,DBLP:conf/iclr/ChenLYZSL20,DBLP:conf/iclr/ParisottoMS0ZK17,DBLP:journals/corr/abs-1808-07980,DBLP:conf/icml/XuZFLB18,DBLP:conf/icml/SantoroHBML18,DBLP:conf/icml/TrivediDWS17,DBLP:conf/icml/AllamanisCKS17}\\
\mbox{[Neuro $\rightarrow$ Symbolic]} & 13 & \cite{DBLP:conf/ijcai/AsaiM20,DBLP:conf/ijcai/DonadelloSG17,DBLP:conf/aaai/AsaiF18,DBLP:conf/nips/CranmerSBXCSH20,DBLP:conf/nips/AlaaS19,DBLP:conf/nips/ZhangSS18,DBLP:conf/iclr/DhingraZBNSC20,DBLP:conf/iclr/MaoGKTW19,DBLP:conf/icml/AmizadehPPHK20,DBLP:conf/icml/LiHHCWZ20,DBLP:conf/icml/ChenBZAGG20,DBLP:conf/icml/Minervini0SGR20,DBLP:conf/icml/MouLLJ17}\\
\mbox{[Neuro[Symbolic]]} & 0  & N/A
\end{tabular}
\caption{Kautz categories paper count. Two paper fit two categories.}\label{tab:kautz}
\end{table}

Categorizations with respect to the 2005 paper are given in Table \ref{tab:survey} and a corresponding paper list in Table \ref{tab:survey-papers}. A number of interesting observations can be made, and we go through each of them in the following.

\begin{table}
\begin{tabular}{llrrr}
& dimension & (a) & (b) & N/A \\
\hline
Interrelation & integrated (a) vs. hybrid (b) & 43 & 0 & 0\\
& neuronal (a) vs. connectionist (b) & 0 & 43 & 0\\
& local (a) vs. distributed (b) & 2 & 42 &  0\\
& standard (a) vs. nonstandard (b) & 43 & 0 &  0\\
\hline
Language & symbolic (a) vs. logical (b) & 21 & 24 & 0\\
& propositional (a) vs. first-order (b) & 3 & 22 & 18 \\
\hline
Usage & extraction (a) vs. representation (b) & 6 & 37 & 3\\
& learning (a) vs. reasoning (b) & 19 & 29 & 0 
\end{tabular}
\caption{2005 survey dimensions paper count. Papers may cover both dimensions. N/A was used if the situation seemed unclear or not applicable.}
\label{tab:survey}
\end{table}

\begin{table}
\begin{tabular}{rl}
dimension & papers \\
\hline
integrated & \cite{DBLP:conf/ijcai/AsaiM20, DBLP:conf/ijcai/YangLLG18,DBLP:conf/ijcai/DonadelloSG17,DBLP:conf/ijcai/XiaoDG17,DBLP:conf/ijcai/PenningGLM11,DBLP:conf/aaai/DemeterD20,DBLP:conf/aaai/LyuYLG19,DBLP:conf/aaai/AsaiF18,DBLP:conf/aaai/CaoLX16,DBLP:conf/nips/Dang-Nhu20,DBLP:conf/nips/ChenLYSZ20,DBLP:conf/nips/CranmerSBXCSH20,DBLP:conf/nips/JiangA20,DBLP:conf/nips/XieXMKS19,DBLP:conf/nips/AlaaS19,DBLP:conf/nips/Yi0G0KT18,DBLP:conf/nips/LiangHZLX18,DBLP:conf/nips/ZhangSS18,DBLP:conf/nips/ManhaeveDKDR18,DBLP:conf/iclr/DhingraZBNSC20,DBLP:conf/iclr/ChenLYZSL20,DBLP:conf/iclr/CohenSHS20,DBLP:conf/iclr/MaoGKTW19,DBLP:conf/iclr/ArabshahiSA18,DBLP:conf/iclr/ParisottoMS0ZK17,DBLP:journals/corr/abs-1808-07980,DBLP:conf/icml/AmizadehPPHK20,DBLP:conf/icml/GargBM20,DBLP:conf/icml/LiHHCWZ20,DBLP:conf/icml/ChenBZAGG20,DBLP:conf/icml/Minervini0SGR20,DBLP:conf/icml/TeruDH20,DBLP:conf/icml/VedantamDLRBP19,DBLP:conf/icml/YoungBN19,DBLP:conf/icml/WangDWK19,DBLP:conf/icml/FischerBDGZV19,DBLP:conf/icml/JiangL19,DBLP:conf/icml/XiongMS16,DBLP:conf/icml/XuZFLB18,DBLP:conf/icml/SantoroHBML18,DBLP:conf/icml/MouLLJ17,DBLP:conf/icml/TrivediDWS17,DBLP:conf/icml/AllamanisCKS17}\\
connectionist & \cite{DBLP:conf/ijcai/AsaiM20, DBLP:conf/ijcai/YangLLG18,DBLP:conf/ijcai/DonadelloSG17,DBLP:conf/ijcai/XiaoDG17,DBLP:conf/ijcai/PenningGLM11,DBLP:conf/aaai/DemeterD20,DBLP:conf/aaai/LyuYLG19,DBLP:conf/aaai/AsaiF18,DBLP:conf/aaai/CaoLX16,DBLP:conf/nips/Dang-Nhu20,DBLP:conf/nips/ChenLYSZ20,DBLP:conf/nips/CranmerSBXCSH20,DBLP:conf/nips/JiangA20,DBLP:conf/nips/XieXMKS19,DBLP:conf/nips/AlaaS19,DBLP:conf/nips/Yi0G0KT18,DBLP:conf/nips/LiangHZLX18,DBLP:conf/nips/ZhangSS18,DBLP:conf/nips/ManhaeveDKDR18,DBLP:conf/iclr/DhingraZBNSC20,DBLP:conf/iclr/ChenLYZSL20,DBLP:conf/iclr/CohenSHS20,DBLP:conf/iclr/MaoGKTW19,DBLP:conf/iclr/ArabshahiSA18,DBLP:conf/iclr/ParisottoMS0ZK17,DBLP:journals/corr/abs-1808-07980,DBLP:conf/icml/AmizadehPPHK20,DBLP:conf/icml/GargBM20,DBLP:conf/icml/LiHHCWZ20,DBLP:conf/icml/ChenBZAGG20,DBLP:conf/icml/Minervini0SGR20,DBLP:conf/icml/TeruDH20,DBLP:conf/icml/VedantamDLRBP19,DBLP:conf/icml/YoungBN19,DBLP:conf/icml/WangDWK19,DBLP:conf/icml/FischerBDGZV19,DBLP:conf/icml/JiangL19,DBLP:conf/icml/XiongMS16,DBLP:conf/icml/XuZFLB18,DBLP:conf/icml/SantoroHBML18,DBLP:conf/icml/MouLLJ17,DBLP:conf/icml/TrivediDWS17,DBLP:conf/icml/AllamanisCKS17}\\
local  & \cite{DBLP:conf/ijcai/PenningGLM11,DBLP:conf/nips/AlaaS19}\\
distributed  & \cite{DBLP:conf/ijcai/AsaiM20, DBLP:conf/ijcai/YangLLG18,DBLP:conf/ijcai/DonadelloSG17,DBLP:conf/ijcai/XiaoDG17,DBLP:conf/ijcai/PenningGLM11,DBLP:conf/aaai/DemeterD20,DBLP:conf/aaai/LyuYLG19,DBLP:conf/aaai/AsaiF18,DBLP:conf/aaai/CaoLX16,DBLP:conf/nips/Dang-Nhu20,DBLP:conf/nips/ChenLYSZ20,DBLP:conf/nips/CranmerSBXCSH20,DBLP:conf/nips/JiangA20,DBLP:conf/nips/XieXMKS19,DBLP:conf/nips/Yi0G0KT18,DBLP:conf/nips/LiangHZLX18,DBLP:conf/nips/ZhangSS18,DBLP:conf/nips/ManhaeveDKDR18,DBLP:conf/iclr/DhingraZBNSC20,DBLP:conf/iclr/ChenLYZSL20,DBLP:conf/iclr/CohenSHS20,DBLP:conf/iclr/MaoGKTW19,DBLP:conf/iclr/ArabshahiSA18,DBLP:conf/iclr/ParisottoMS0ZK17,DBLP:journals/corr/abs-1808-07980,DBLP:conf/icml/AmizadehPPHK20,DBLP:conf/icml/GargBM20,DBLP:conf/icml/LiHHCWZ20,DBLP:conf/icml/ChenBZAGG20,DBLP:conf/icml/Minervini0SGR20,DBLP:conf/icml/TeruDH20,DBLP:conf/icml/VedantamDLRBP19,DBLP:conf/icml/YoungBN19,DBLP:conf/icml/WangDWK19,DBLP:conf/icml/FischerBDGZV19,DBLP:conf/icml/JiangL19,DBLP:conf/icml/XiongMS16,DBLP:conf/icml/XuZFLB18,DBLP:conf/icml/SantoroHBML18,DBLP:conf/icml/MouLLJ17,DBLP:conf/icml/TrivediDWS17,DBLP:conf/icml/AllamanisCKS17}\\
standard & \cite{DBLP:conf/ijcai/AsaiM20, DBLP:conf/ijcai/YangLLG18,DBLP:conf/ijcai/DonadelloSG17,DBLP:conf/ijcai/XiaoDG17,DBLP:conf/ijcai/PenningGLM11,DBLP:conf/aaai/DemeterD20,DBLP:conf/aaai/LyuYLG19,DBLP:conf/aaai/AsaiF18,DBLP:conf/aaai/CaoLX16,DBLP:conf/nips/Dang-Nhu20,DBLP:conf/nips/ChenLYSZ20,DBLP:conf/nips/CranmerSBXCSH20,DBLP:conf/nips/JiangA20,DBLP:conf/nips/XieXMKS19,DBLP:conf/nips/AlaaS19,DBLP:conf/nips/Yi0G0KT18,DBLP:conf/nips/LiangHZLX18,DBLP:conf/nips/ZhangSS18,DBLP:conf/nips/ManhaeveDKDR18,DBLP:conf/iclr/DhingraZBNSC20,DBLP:conf/iclr/ChenLYZSL20,DBLP:conf/iclr/CohenSHS20,DBLP:conf/iclr/MaoGKTW19,DBLP:conf/iclr/ArabshahiSA18,DBLP:conf/iclr/ParisottoMS0ZK17,DBLP:journals/corr/abs-1808-07980,DBLP:conf/icml/AmizadehPPHK20,DBLP:conf/icml/GargBM20,DBLP:conf/icml/LiHHCWZ20,DBLP:conf/icml/ChenBZAGG20,DBLP:conf/icml/Minervini0SGR20,DBLP:conf/icml/TeruDH20,DBLP:conf/icml/VedantamDLRBP19,DBLP:conf/icml/YoungBN19,DBLP:conf/icml/WangDWK19,DBLP:conf/icml/FischerBDGZV19,DBLP:conf/icml/JiangL19,DBLP:conf/icml/XiongMS16,DBLP:conf/icml/XuZFLB18,DBLP:conf/icml/SantoroHBML18,DBLP:conf/icml/MouLLJ17,DBLP:conf/icml/TrivediDWS17,DBLP:conf/icml/AllamanisCKS17}\\
symbolic & \cite{DBLP:conf/ijcai/AsaiM20, DBLP:conf/ijcai/YangLLG18,DBLP:conf/ijcai/XiaoDG17,DBLP:conf/ijcai/PenningGLM11,DBLP:conf/aaai/DemeterD20,DBLP:conf/aaai/LyuYLG19,DBLP:conf/aaai/AsaiF18,DBLP:conf/aaai/CaoLX16,DBLP:conf/nips/Dang-Nhu20,DBLP:conf/nips/JiangA20,DBLP:conf/nips/AlaaS19,DBLP:conf/nips/ZhangSS18,DBLP:conf/iclr/ArabshahiSA18,DBLP:conf/icml/AmizadehPPHK20,DBLP:conf/icml/GargBM20,DBLP:conf/icml/LiHHCWZ20,DBLP:conf/icml/ChenBZAGG20,DBLP:conf/icml/Minervini0SGR20,DBLP:conf/icml/SantoroHBML18,DBLP:conf/icml/MouLLJ17,DBLP:conf/icml/TrivediDWS17}\\
logical  & \cite{DBLP:conf/ijcai/DonadelloSG17,DBLP:conf/ijcai/PenningGLM11,DBLP:conf/nips/ChenLYSZ20,DBLP:conf/nips/CranmerSBXCSH20,DBLP:conf/nips/XieXMKS19,DBLP:conf/nips/Yi0G0KT18,DBLP:conf/nips/LiangHZLX18,DBLP:conf/nips/ManhaeveDKDR18,DBLP:conf/iclr/DhingraZBNSC20,DBLP:conf/iclr/ChenLYZSL20,DBLP:conf/iclr/CohenSHS20,DBLP:conf/iclr/MaoGKTW19,DBLP:conf/iclr/ParisottoMS0ZK17,DBLP:journals/corr/abs-1808-07980,DBLP:conf/icml/AmizadehPPHK20,DBLP:conf/icml/TeruDH20,DBLP:conf/icml/VedantamDLRBP19,DBLP:conf/icml/YoungBN19,DBLP:conf/icml/WangDWK19,DBLP:conf/icml/FischerBDGZV19,DBLP:conf/icml/JiangL19,DBLP:conf/icml/XiongMS16,DBLP:conf/icml/XuZFLB18,DBLP:conf/icml/AllamanisCKS17}\\
propositional & \cite{DBLP:conf/ijcai/PenningGLM11,DBLP:conf/nips/XieXMKS19,DBLP:conf/icml/AllamanisCKS17}\\
first-order  & \cite{DBLP:conf/nips/ChenLYSZ20,DBLP:conf/nips/CranmerSBXCSH20,DBLP:conf/nips/Yi0G0KT18,DBLP:conf/nips/LiangHZLX18,DBLP:conf/nips/ManhaeveDKDR18,DBLP:conf/iclr/DhingraZBNSC20,DBLP:conf/iclr/ChenLYZSL20,DBLP:conf/iclr/CohenSHS20,DBLP:conf/iclr/MaoGKTW19,DBLP:conf/iclr/ParisottoMS0ZK17,DBLP:journals/corr/abs-1808-07980,DBLP:conf/icml/AmizadehPPHK20,DBLP:conf/icml/TeruDH20,DBLP:conf/icml/VedantamDLRBP19,DBLP:conf/icml/YoungBN19,DBLP:conf/icml/WangDWK19,DBLP:conf/icml/FischerBDGZV19,DBLP:conf/icml/JiangL19,DBLP:conf/icml/XiongMS16,DBLP:conf/icml/XuZFLB18}\\
extraction  & \cite{DBLP:conf/ijcai/AsaiM20,DBLP:conf/ijcai/PenningGLM11,DBLP:conf/aaai/AsaiF18,DBLP:conf/nips/Dang-Nhu20,DBLP:conf/icml/AmizadehPPHK20,DBLP:conf/iclr/ChenLYZSL20}\\
representation  & \cite{DBLP:conf/ijcai/DonadelloSG17,DBLP:conf/ijcai/PenningGLM11,DBLP:conf/aaai/CaoLX16,DBLP:conf/nips/ChenLYSZ20,DBLP:conf/nips/CranmerSBXCSH20,DBLP:conf/nips/JiangA20,DBLP:conf/nips/XieXMKS19,DBLP:conf/nips/AlaaS19,DBLP:conf/nips/Yi0G0KT18,DBLP:conf/nips/LiangHZLX18,DBLP:conf/nips/ZhangSS18,DBLP:conf/nips/ManhaeveDKDR18,DBLP:conf/iclr/DhingraZBNSC20,DBLP:conf/iclr/ChenLYZSL20,DBLP:conf/iclr/CohenSHS20,DBLP:conf/iclr/MaoGKTW19,DBLP:conf/iclr/ArabshahiSA18,DBLP:conf/iclr/ParisottoMS0ZK17,DBLP:journals/corr/abs-1808-07980,DBLP:conf/icml/AmizadehPPHK20,DBLP:conf/icml/GargBM20,DBLP:conf/icml/LiHHCWZ20,DBLP:conf/icml/ChenBZAGG20,DBLP:conf/icml/Minervini0SGR20,DBLP:conf/icml/TeruDH20,DBLP:conf/icml/VedantamDLRBP19,DBLP:conf/icml/YoungBN19,DBLP:conf/icml/WangDWK19,DBLP:conf/icml/FischerBDGZV19,DBLP:conf/icml/JiangL19,DBLP:conf/icml/XiongMS16,DBLP:conf/icml/XuZFLB18,DBLP:conf/icml/SantoroHBML18,DBLP:conf/icml/MouLLJ17,DBLP:conf/icml/TrivediDWS17,DBLP:conf/icml/AllamanisCKS17} \\
learning  & \cite{DBLP:conf/ijcai/AsaiM20,DBLP:conf/ijcai/YangLLG18,DBLP:conf/ijcai/DonadelloSG17,DBLP:conf/ijcai/XiaoDG17,DBLP:conf/ijcai/PenningGLM11,DBLP:conf/aaai/DemeterD20,DBLP:conf/aaai/LyuYLG19,DBLP:conf/aaai/AsaiF18,DBLP:conf/aaai/CaoLX16,DBLP:conf/nips/Dang-Nhu20,DBLP:conf/nips/ChenLYSZ20,DBLP:conf/nips/JiangA20,DBLP:journals/corr/abs-1808-07980,DBLP:conf/icml/GargBM20,DBLP:conf/icml/VedantamDLRBP19,DBLP:conf/icml/YoungBN19,DBLP:conf/icml/JiangL19,DBLP:conf/icml/SantoroHBML18,DBLP:conf/icml/TrivediDWS17}\\
reasoning  & \cite{DBLP:conf/ijcai/DonadelloSG17,DBLP:conf/ijcai/PenningGLM11,DBLP:conf/nips/Dang-Nhu20,DBLP:conf/nips/CranmerSBXCSH20,DBLP:conf/nips/XieXMKS19,DBLP:conf/nips/AlaaS19,DBLP:conf/nips/Yi0G0KT18,DBLP:conf/nips/LiangHZLX18,DBLP:conf/nips/ZhangSS18,DBLP:conf/nips/ManhaeveDKDR18,DBLP:conf/iclr/DhingraZBNSC20,DBLP:conf/iclr/ChenLYZSL20,DBLP:conf/iclr/CohenSHS20,DBLP:conf/iclr/MaoGKTW19,DBLP:conf/iclr/ArabshahiSA18,DBLP:conf/iclr/ParisottoMS0ZK17,DBLP:journals/corr/abs-1808-07980,DBLP:conf/icml/AmizadehPPHK20,DBLP:conf/icml/LiHHCWZ20,DBLP:conf/icml/ChenBZAGG20,DBLP:conf/icml/Minervini0SGR20,DBLP:conf/icml/TeruDH20,DBLP:conf/icml/VedantamDLRBP19,DBLP:conf/icml/WangDWK19,DBLP:conf/icml/FischerBDGZV19,DBLP:conf/icml/XiongMS16,DBLP:conf/icml/XuZFLB18,DBLP:conf/icml/MouLLJ17,DBLP:conf/icml/AllamanisCKS17}
\end{tabular}
\caption{2005 survey dimensions paper classification. Dimensions with no entries have been omitted.}\label{tab:survey-papers}
\end{table}

Regarding the integrated vs. hybrid dimension, we are taking a similar tack as the 2005 survey: this is mostly for delineation, i.e. we focus on integrated approaches.

Regarding the neuronal vs. connectionist dimension, it is noteworthy but perhaps unsurprising that there were no papers with emphasis on biological plausibility: Much of the current rise of NeSy AI as a topic is fueled by deep learning with a heavy emphasis on desireable computational functionality.  The 2005 survey, however, shows a much more balanced picture, and indeed several early works, such as \cite{DBLP:journals/apin/Shastri99}, put much more emphasis on biological plausibility. However, it is also conceivable that such papers may not be considered as interesting by typical top tier AI conference reviewers, and thus may seek publication outlets elsewhere. 

Similar considerations may apply on both the local vs. distributed and the standard vs. nonstandard network architecture dimension. The recent heavy influence of deep learning, as well as our restriction to top tier AI conferences may explain the lopsided counts: architectures are more-or-less standard deep learning architectures, and in these, knowledge is usually encoded in a distributed fashion using some kind of embeddings in multi-dimensional Euclidean space. In the 2005 survey, nonstandard architectures and localist representations featured much more prominently, and the previously cited \cite{DBLP:journals/apin/Shastri99} is also an example for such work.

The counts on the symbolic vs. logical dimension mark another departure from the situation described in the 2005 paper, which indicated a much stronger emphasis on formal logic in NeSy AI,\footnote{Symbolic but non-logical aspects were of course also relevant, see e.g. \cite{nesy-book-07} where seven out of twelve chapters did not touch on logical aspects.} as opposed to symbolic aspects. 

The lopsided count for first-order logic as opposed to propositional logic illustrates another shift in the subfield. It was  discussed in the 2005 survey that a majority of NeSy AI work appears to focus on propositional aspects, it was previously understood that developing NeSy AI solutions is much harder when dealing with logics that are more expressive than propositional logics. John McCarthy indeed used the term "propositional fixation" of artificial neural networks to point this out, as early as 1988 \cite{mccarthy-propfix}. That first-order logics beyond propositional play a more prominent role now is an indicator not only of new and enhanced capabilities, it also means that it is now possible to utilize logics that are of increasing relevance to the Knowledge Representation and Reasoning subfield of AI. This is an excellent and promising development. The N/A papers in this row in Table \ref{tab:survey} indicate paper that did not involve logic at all, and instead used symbolic information that is not based on formal logic.

On the extraction vs. representation dimension, we notice that explicit work on extracting symbolic information from trained neural networks is less of an emphasis. This makes sense in the light of Kautz' categories above, where the first three categories would usually not involve an extraction aspect, and the fourth also may not. Knowledge extraction was a more prominent and emphasized topic in the past.

On the learning vs. reasoning dimension, we see a rather balanced count, which indicates that both aspects are not only important, as is to be expected, but can actually be done. In the past, both learning of first-order logic and reasoning with it was particularly difficult for systems based on artificial neural networks, but it appears that large strides have been made since the advent of deep learning and the growing understanding that these aspects are of importance for advancing the AI field.

Out of interest, we also looked at the selected 43 papers from a perspective of some possible benefits that may arise from integrating neural and symbolic systems, as indicated in the introduction. The following seemed particularly pertinent to us:

\begin{description}
\item[\mbox{Out of distribution handling:}] Whether the designed system is able to extrapolate from training input data by means of utilizing symbolic means such as background knowledge, and thus is able to handle scenarios that are significantly different from the training input. 
\item[\mbox{Intepretability:}] Whether the fact that the system is neuro-symbolic is instrumental to enhanced interpretability of system behavior or outputs, e.g. in terms of making system decisions more transparent and explainable for a human user.
\item[\mbox{Error recovery:}] Whether the designed system makes use of its neuro-symbolic design in order to recover more easily from erroneous decisions or outputs.
\item[\mbox{Learning from small data:}] Whether the fact that the system is neuro-symbolic is instrumental in making the system trainable on less data than you would expect from a system that is not neuro-symbolic.
\end{description}

As for the previous categorizations, decisions how to classify each paper were often not clear-cut. We usually went with the authors' own positioning, if there was one. The resulting counts can be found in Table \ref{tab:features}. We see that interpretability aspects are most common. Only few papers address out of distribution issues or learning from small data, and there is hardly any work related to error recovery.

\begin{table}
\begin{tabular}{rll}
feature & number of papers & papers \\
\hline
Out of distribution handling & 8 & \cite{DBLP:conf/ijcai/DonadelloSG17,DBLP:conf/ijcai/PenningGLM11,DBLP:conf/aaai/DemeterD20,DBLP:conf/nips/Dang-Nhu20,DBLP:conf/nips/ChenLYSZ20,DBLP:conf/nips/LiangHZLX18,DBLP:conf/iclr/MaoGKTW19,DBLP:conf/icml/TeruDH20}\\
Interpretability & 25 & \cite{DBLP:conf/ijcai/YangLLG18,DBLP:conf/ijcai/PenningGLM11,DBLP:conf/aaai/LyuYLG19,DBLP:conf/nips/Dang-Nhu20,DBLP:conf/nips/ChenLYSZ20,DBLP:conf/nips/CranmerSBXCSH20,DBLP:conf/nips/XieXMKS19,DBLP:conf/nips/AlaaS19,DBLP:conf/nips/Yi0G0KT18,DBLP:conf/nips/LiangHZLX18,DBLP:conf/nips/ZhangSS18,DBLP:conf/nips/ManhaeveDKDR18,DBLP:conf/iclr/ChenLYZSL20,DBLP:conf/iclr/CohenSHS20,DBLP:conf/iclr/MaoGKTW19,DBLP:conf/iclr/ParisottoMS0ZK17,DBLP:conf/icml/Minervini0SGR20,DBLP:conf/icml/VedantamDLRBP19,DBLP:conf/icml/YoungBN19,DBLP:conf/icml/WangDWK19,DBLP:conf/icml/FischerBDGZV19,DBLP:conf/icml/JiangL19,DBLP:conf/icml/XiongMS16,DBLP:conf/icml/XuZFLB18,DBLP:conf/icml/AllamanisCKS17}\\
Error recovery & 2 & \cite{DBLP:conf/nips/ZhangSS18,DBLP:conf/icml/VedantamDLRBP19} \\
Learning from small data & 7 & \cite{DBLP:conf/ijcai/YangLLG18,DBLP:conf/aaai/DemeterD20,DBLP:conf/aaai/LyuYLG19,DBLP:conf/nips/Dang-Nhu20,DBLP:conf/nips/Yi0G0KT18,DBLP:conf/iclr/ParisottoMS0ZK17,DBLP:conf/icml/VedantamDLRBP19} 
\end{tabular}
\caption{Some desirable neuro-symbolic features.}\label{tab:features}
\end{table}

\section{Paths Forward}\label{sec:future}

Like so many recent topic shifts in AI, it appears that the increased attention to NeSy AI is mostly due to the rise of deep learning, which has shown tremendous success even in areas which previously were mostly in the domain of symbolic approaches, such as Chess or Go game engines. There seems to be a reasonable expectation that deep learning solutions may be much more suitable to address the subsymbolic-symbolic gap than previous connectionist machine learning technology.

However, despite these advances and promises, it seems imperative to explicitly explore and understand the capabilities and limitations of deep learning based symbolic manipulation, as a basis for further progress on combining neural and symbolic aspects in a best of both worlds fashion. Indeed, a systematic exploration of the extent to which deep learning systems can learn straightforward and well-understood symbol manipulation tasks would shed significant light on this question. Possible concrete symbol manipulation tasks for study can be found all over AI and computer science, such as term rewriting, list, tree and graph manipulations, executing formal grammars, elementary algebra, logical deduction. In-depth studies of these from a deep learning perspective would provide systems with elementary capabilities that can then be composed for more complex solutions, or used as modules in larger AI systems \cite{HarmelenT19}. 

At the same time, it also appears to be important to contrast recent NeSy AI publications with the much larger vision of the subfield. Indeed, the current promise of NeSy AI lies in a favorable combination or integration of deep learning with symbolic AI approaches from the subfield of Knowledge Representation and Reasoning, where complex formal logics dominate. As we have seen in Section \ref{sec:past}, current NeSy AI publications at top conferences involve complex logics to a much larger extent than was the case two decades ago, however much of the work is still mostly on the non-logical symbolic level, and desirable features which conceptually should arise out of a systematic use of neuro-symbolic approaches, such as those listed in Table \ref{tab:features}, are often still not met. Whether it is possible, or to what extent, to achieve a much stronger integration of complex formal logics and deep learning, including best-of-both-worlds features, is of course currently not known, and is in fact in itself a fundamental research question that remains to be addressed.

We thus posit that more emphasis is needed, in the immediate future, on deepening the logical aspects in NeSy AI research even further, and to work towards a systematic understanding and toolbox for utilizing complex logics in this context. In particular, it would be helpful for advancing the field to conduct experiments with different levels of sophistication in the logical aspects. For example, systems that utilize "flat" annotations (metadata tags that are simply keywords) are essentially operating on the logical level of "facts" only. We can then ask the question whether such a system can be improved by using, say, a class hierarchy of annotations (such as schema.org), which corresponds to making use of a knowledge base with simple logical  implications (i.e., subclass relationships). If the answer to this is affermative, then more complex logical background knowledge can be attempted to be leveraged. The Semantic Web field of research \cite{Hitzler21} for example, provides a whole hierarchy of logics on metadata \cite{FOST} that can be utilized, at the level at which a deep learning system can ingest them for added value, and can thus be an excellent playground for developing fundamental NeSy AI capabilities \cite{HitzlerBES20}.

An example for a currently emerging line of research compatible with the thoughts just laid out, is that of so-called \emph{deep deductive reasoners} \cite{EEBH21-apin}, which are deep learning systems trained to perform deductive reasoning over formal logics. The deductive reasoning task can be framed as classification (a statement is or is not a logical consequence of a given theory) or as regression (produce all logical consequences under some finiteness constraints). Different deep learning architectures have already been investigated, on logics that range from very straightforward ones like RDF \cite{FOST} to first-order predicate logic. While the body of work is still relatively small, it is obvious from the published results \cite{DBLP:journals/corr/abs-1808-07980,DBLP:journals/semweb/MakniH19,DBLP:conf/aaaiss/BianchiH19,EbrahimiSBXEDKH21,DBLP:conf/aaaiss/EberhartEZSH20} that deductive reasoning is a very hard task for deep learning, and with increasing hardness and scalability issues as logics become more complex. 

\section{Conclusions}\label{sec:conc}

The recent rise in publications, events, and public addresses on NeSy AI indicates that there is increasing awareness of the importance of the subfield in the mainstream AI community. Arguably, this is also happening at the inflection point in time when we are slowly beginning to explore and understand the inherent limitations of pure deep learning approaches. The use of additional background knowledge is a natural path to attempt to further improve deep learning systems, and much of this line of work falls into the NeSy AI theme. 

The broad consensus seems to be that combining neural and symbolic approaches in the sense of NeSy AI is at least a path forward to much stronger AI systems. Arguably, it may also turn out to be a major stepping stone towards human-level artificial intelligence.







\bibliographystyle{ios1}           
\bibliography{bibliography}        

%

\end{document}